\documentclass[conference]{IEEEtran}
\IEEEoverridecommandlockouts

\usepackage{cite}
\usepackage{amsmath,amssymb,amsfonts}
\usepackage{algorithmic}
\usepackage{graphicx}
\usepackage{textcomp}
\usepackage{xcolor}
\usepackage{booktabs}
\usepackage{multirow}
\usepackage{url}
\usepackage{enumitem}
\usepackage{balance}

\def\BibTeX{{\rm B\kern-.05em{\sc i\kern-.025em b}\kern-.08em
    T\kern-.1667em\lower.7ex\hbox{E}\kern-.125emX}}

\begin{document}

\setlength{\parskip}{0pt}
\makeatletter
\def\section{\@startsection{section}{1}{\z@}{6pt}{4pt}%
{\normalfont\normalsize\centering\scshape}}%
\def\subsection{\@startsection{subsection}{2}{\z@}{4pt}{3pt}%
{\normalfont\normalsize\itshape}}%
\def\subsubsection{\@startsection{subsubsection}{3}{\z@}{3pt}{2pt}%
{\normalfont\normalsize\itshape}}%
\makeatother
\setlist{nosep, topsep=0pt}
\setlength{\floatsep}{4pt}
\setlength{\textfloatsep}{6pt}
\setlength{\intextsep}{4pt}
\setlength{\abovedisplayskip}{4pt}
\setlength{\belowdisplayskip}{4pt}
\setlength{\abovedisplayshortskip}{2pt}
\setlength{\belowdisplayshortskip}{2pt}

\title{Global Ranks Survive, Selected Heads Shift: BOS-Sink Topology
under 4-bit Weight-Only Quantization\thanks{\copyright~2026 IEEE. Personal use of this material is permitted. Permission from IEEE must be obtained for all other uses, in any current or future media, including reprinting/republishing this material for advertising or promotional purposes, creating new collective works, for resale or redistribution to servers or lists, or reuse of any copyrighted component of this work in other works. Accepted for publication at IEEE IECON 2026.}}

\author{
\IEEEauthorblockN{Kuanlin Chen}
\IEEEauthorblockA{
\textit{Independent Researcher}\\
Taoyuan, Taiwan\\
dipper13579@gmail.com\\
ORCID: 0009-0006-3879-4565
}
\and
\IEEEauthorblockN{Chen-Wei Kuo}
\IEEEauthorblockA{
\textit{National Tsing Hua University}\\
Hsinchu, Taiwan\\
ck3294@nyu.edu
}
\and
\IEEEauthorblockN{Cheng-En Ou}
\IEEEauthorblockA{
\textit{Independent Researcher}\\
Taoyuan, Taiwan\\
garyandevan3@gmail.com
}
}

\maketitle

\begin{abstract}
Sink-aware deployment may identify important first-token attention
heads before a model is quantized, then reuse that map at the edge.
We test when this shortcut is safe for 4-bit NF4 weight-only
post-training quantization (PTQ). Our Sink Topology Consistency (STC)
metrics separate global rank preservation, top-$k$ set overlap, and
layerwise sink-mass shift, and distinguish per-input sensitivity from
calibration-map transfer. Across Qwen2.5-0.5B, Qwen2.5-1.5B, and
Llama-3.2-1B, global bf16--4-bit ranks remain high at 4,096 tokens
($\rho_s\geq0.980$), yet top-$k$ Jaccard overlap is only
$0.619$--$0.793$, corresponding to $76.5$--$88.5\%$ membership
retention. The global statistic also masks local failures: terminal
Qwen layers shift by $6.2$--$7.9\times$ their model means, whereas
Llama-3.2-1B shows low, nearly uniform drift. Under a C4-to-LongBench
shift, cross-domain overlap degrades more than the within-domain
precision comparison for both Qwen models, but not for Llama-3.2-1B.
Matched-domain 4-bit recalibration reaches 90\% of a split-half
stability plateau at the smallest tested $n{=}8$ for both Qwen models
and $n{=}32$ for Llama-3.2-1B, though not as a sharp threshold; for
the two Qwen models, updating only selected layers does not reach the
full-map stability criterion.
On Jetson Orin NX, the 16-sample workload takes seconds for the two
models with valid on-device sink measurements. The practical message
is precise: global rankings often transfer, but discrete head sets,
layer-local policies, and cross-domain calibration should be
revalidated after quantization.
\end{abstract}


\section{Introduction}
\label{sec:intro}

Small language models enable document retrieval, fault diagnosis,
and procedure checking on resource-constrained edge hardware. A
1--3B parameter model occupies 2--6\,GB in bf16; 4-bit weight
quantization cuts the weight footprint by roughly $4\times$ with
manageable accuracy loss~\cite{frantar2022gptq,lin2024awq}. A natural
deployment shortcut is therefore to identify useful attention
structure on a higher-resource workstation and reuse it after
quantization. This paper asks when that shortcut preserves the
head-level first-token sink map.

A growing family of inference-time methods exploits the attention
sink phenomenon --- the tendency of autoregressive language models
to assign high attention weight to the first token position
regardless of its semantic content~\cite{xiao2024streamingllm}.
StreamingLLM retains initial-token KV states for stable
long-context generation~\cite{xiao2024streamingllm}; CushionCache
inserts prefix sink tokens to reduce activation outliers that
impede quantization~\cite{son2024prefixing}; SINKTRACK injects key
context features into the BOS representation to anchor against
context forgetting~\cite{sinktrack2026}. These methods operate at
different levels --- token retention, prefix insertion, and BOS
anchoring --- and do not all select sink heads explicitly. They
nevertheless motivate a narrower deployment question: if a pipeline
calibrates a head-level first-token sink map before compression,
which parts remain valid after 4-bit weight-only quantization?

The interaction between PTQ and sink structure is not theoretically
neutral. Quantization introduces weight rounding errors that are
non-uniform across layers and heads, and because attention weights
come from a softmax over quantization-perturbed logits, even small
logit perturbations can change whether a head crosses a
sink-dominance criterion --- altering the composition of the
sink-dominant set. The first token's key vector also has
anomalously small L2-norm~\cite{gu2025attention}, so quantization
error on its key projection may have a disproportionate relative
effect. This makes head-level sink stability under weight-only PTQ
an open empirical question.

We contribute: (i) a \textbf{calibration-transfer formulation} for
bf16-to-4-bit sink maps; (ii) the \textbf{STC metric family}, which
separates per-input and calibration-level set overlap, global rank
preservation, and layerwise sink-mass shift; (iii) an empirical study
across three small open-weight models, three context lengths, and a
cross-domain probe; and (iv) \textbf{deployment guidance} that
distinguishes reusable global rankings from head sets and local
structures that require revalidation.

\section{Related Work}

Attention sink emerges from softmax normalization, which forces
attention weights to sum to one even when no semantically relevant
token exists~\cite{gu2025attention}: the first token acts as an
implicit key bias, absorbing surplus attention probability without
contributing to value
computation~\cite{gu2025attention,sun2024massive}. It has been
further explained as preventing representational over-mixing in
deep residual networks~\cite{barbero2025llms}, and characterized
geometrically as establishing reference frames in token
space~\cite{ruscio2025geometric}. The inference-time methods built
on it --- StreamingLLM, CushionCache and
SINKTRACK~\cite{xiao2024streamingllm,son2024prefixing,sinktrack2026}
--- are described in Section~\ref{sec:intro}.

Post-training quantization compresses weights to 4-bit or lower
without retraining~\cite{frantar2022gptq,lin2024awq}. The massive
activations co-located with sink tokens~\cite{sun2024massive}
distort the dynamic range of quantization groups; SmoothQuant~\cite{xiao2023smoothquant}
and AWQ~\cite{lin2024awq} mitigate this at the activation level,
but target task accuracy rather than structural properties of
attention. Sink topology is known to be perturbation-sensitive:
dormant sink heads can be reactivated by surgical weight
reinitialization~\cite{guo2024dormant}, and positional encoding
alters which heads become sink-dominant~\cite{ruscio2025geometric}.
Yet no prior study has measured whether weight-only PTQ --- the
dominant edge compression strategy --- preserves the head-level
BOS-sink topology that sink-aware methods rely upon. We address
this gap.

\section{Method}

\subsection{Operational Definition}

All measurements are taken at token index 0, which serves as the
first-token sink position whether or not the tokenizer prepends a
dedicated BOS token: for Llama-3.2-1B index~0 is
\texttt{<|begin\_of\_text|>} (a true BOS token), while for Qwen2.5,
which prepends none by default, it is the first content or
chat-template control token. The first-token sink is observed at
index~0 in both
cases~\cite{xiao2024streamingllm,gu2025attention}. We use
``BOS-sink'' as conventional shorthand for first-token sink
throughout.

\subsection{Sink Score}

The per-sample sink score for layer $l$, head $h$, and input $x$
of length $T$ is:
\begin{equation}
  s_{l,h}(x) = \frac{1}{T-1}\sum_{t=1}^{T-1}\alpha_{l,h}(t\!\to\!0)
  \label{eq:scs}
\end{equation}
where $\alpha_{l,h}(t\!\to\!0)$ is the post-softmax attention weight
from query position $t$ to key position 0 under
\texttt{attn\_implementation="eager"}. The cross-sample mean is:
\begin{equation}
  \bar{s}_{l,h} = \frac{1}{|\mathcal{D}|}\sum_{x\in\mathcal{D}}s_{l,h}(x)
  \label{eq:scs_mean}
\end{equation}

\subsection{STC Metric Family}

We introduce two operationalizations of Sink Topology Consistency
(STC) serving distinct engineering purposes.

\subsubsection{Sample-STC (topology sensitivity)}
Measures how stable sink topology is on individual inputs. Let
$H_k^{\text{bf16}}(x) = \text{TopK}_{(l,h)}\,s_{l,h}^{\text{bf16}}(x)$
and similarly for 4-bit, where $k = \lceil 0.10 \times L H_q \rceil$
($H_q$ = number of query heads). For example, Qwen2.5-1.5B has
$L{=}28$, $H_q{=}12$, giving $k = \lceil 0.10 \times 336 \rceil = 34$
(see Table~\ref{tab:models}).

\begin{align}
  \text{Samp-Set@}k &= \mathbb{E}_x\!\left[
    \frac{|H_k^{\text{bf16}}(x)\cap H_k^{\text{4bit}}(x)|}
         {|H_k^{\text{bf16}}(x)\cup H_k^{\text{4bit}}(x)|}
  \right] \label{eq:samp_set}\\
  \text{Samp-Rank} &= \mathbb{E}_x\!\left[
    \rho_s\!\left(\mathbf{s}^{\text{bf16}}(x),\mathbf{s}^{\text{4bit}}(x)\right)
  \right] \label{eq:samp_rank}\\
  \text{Samp-LS} &= \mathbb{E}_x\!\left[
    \frac{1}{L}\sum_l \mathrm{JS}\!\left(p_l^{\text{bf16}}(x),
    p_l^{\text{4bit}}(x)\right)
  \right] \label{eq:samp_ls}
\end{align}
where $\rho_s$ is Spearman rank correlation over all $LH_q$
head slots and $p_l$ is the per-layer sink-mass distribution
normalized over heads. Throughout, $\mathrm{JS}$ denotes the
Jensen--Shannon \emph{distance} --- the square root of the
Jensen--Shannon divergence in natural-log base, as implemented by
\texttt{scipy.spatial.distance.jensenshannon} --- so
LayerShift takes values in $[0,\sqrt{\ln 2}\,]\approx[0,0.833]$.

\subsubsection{Cal-STC (calibration transfer)}
Measures whether a calibration-derived sink map survives precision
change. Let $\bar{H}_k^{\text{bf16}} = \text{TopK}_{(l,h)}\bar{s}_{l,h}^{\text{bf16}}$.

\begin{align}
  \text{Cal-Set@}k &= \frac{|\bar{H}_k^{\text{bf16}}\cap
    \bar{H}_k^{\text{4bit}}|}{|\bar{H}_k^{\text{bf16}}\cup
    \bar{H}_k^{\text{4bit}}|} \label{eq:cal_set}\\
  \text{Cal-Rank} &= \rho_s\!\left(\bar{\mathbf{s}}^{\text{bf16}},
    \bar{\mathbf{s}}^{\text{4bit}}\right) \label{eq:cal_rank}\\
  \text{Cal-LS} &= \frac{1}{L}\sum_l
    \mathrm{JS}\!\left(\bar{p}_l^{\text{bf16}},
    \bar{p}_l^{\text{4bit}}\right) \label{eq:cal_ls}
\end{align}

\textbf{One-line distinction:} Sample-STC answers ``is any single
input's sink map stable?''; Cal-STC answers ``does my offline
calibration survive quantization?''

\subsubsection{Null baseline}
For two independently drawn $k$-subsets of the $LH_q$ head slots,
the ratio of expected intersection to expected union is
$k/(2LH_q-k)$. This gives $0.0533$ for both Qwen models
($k{=}34$, $LH_q{=}336$) and $0.0535$ for Llama-3.2-1B
($k{=}52$, $LH_q{=}512$); we report this closed form as the null
baseline. We additionally draw 1,000 independent subset pairs per
model, which yields a Monte Carlo 95\% interval of $[0.000,0.115]$
for the Qwen models and $[0.020,0.106]$ for Llama-3.2-1B.

\subsection{Quantization Protocol}

All main experiments are conducted on a single NVIDIA RTX 3060\,Ti
(8\,GB VRAM) under Windows with CUDA 12.1. Jetson validation uses
an NVIDIA Jetson Orin NX 8\,GB with JetPack 6.x.

The reported desktop runs use one environment (Python 3.11.14,
PyTorch 2.5.1+cu121, HuggingFace Transformers 5.3.0, and
bitsandbytes 0.49.2). Within each paired comparison, the checkpoint,
input tokens, teacher-forced path (\texttt{use\_cache=False}), and
eager-attention backend are fixed. Baseline precision is
bfloat16 (bf16); quantized precision is 4-bit weight-only NF4 via
bitsandbytes with parameters \texttt{bnb\_4bit\_quant\_type="nf4"},
\texttt{bnb\_4bit\_compute\_dtype=torch.float16}, and
\texttt{bnb\_4bit\_use\_double\_quant=False} (double quantization
disabled to isolate standard NF4 behavior).
Eager attention (\texttt{attn\_implementation="eager"}) is required
for \texttt{output\_attentions=True}. This avoids an attention-backend
change, but the comparison couples NF4 weights with fp16 compute
against a bf16 baseline; it does not isolate weight rounding alone.
Jetson validation uses transformers 4.57.6, the platform's
available version; a known attention-output format incompatibility
under NF4 + eager attention on arm64 affects Qwen2.5-1.5B hook
measurement at that version (see Table~\ref{tab:jetson}).

Memory-safe measurement uses forward hooks that accumulate only
the BOS-column statistics $(\alpha_{l,h}(t\to0)$ means) per layer,
discarding the full attention tensor immediately; this avoids OOM
at 4,096 tokens on consumer GPUs. All forward passes use batch
size 1.
Hook validity is confirmed by comparing hook-based SCS against
direct \texttt{outputs.attentions} on a 128-token sequence,
against an acceptance threshold of $10^{-3}$; the observed maximum
absolute deviation is below $10^{-6}$ for all tested models.

\subsection{Models}

\begin{table}[ht]
\caption{Model Specifications}
\label{tab:models}
\centering
\begin{tabular}{lccccc}
\toprule
Model & Params & Layers & $H_q$ & $H_{kv}$ & $k$ \\
\midrule
Qwen2.5-0.5B & 500M & 24 & 14 & 2  & 34 \\
Qwen2.5-1.5B & 1.5B & 28 & 12 & 2  & 34 \\
Llama-3.2-1B & 1.0B & 16 & 32 & 8  & 52 \\
\bottomrule
\end{tabular}
\end{table}

All models use RoPE positional encoding and GQA, representative
of current edge deployment targets. Instruct variants are used
throughout. In GQA models, STC is defined over query heads; the
returned attention tensors reflect query-head distributions after
KV expansion. Note that the query-to-KV head ratio $H_q/H_{kv}$
is $7$ for Qwen2.5-0.5B, $6$ for Qwen2.5-1.5B and $4$ for
Llama-3.2-1B: Llama-3.2-1B has by far the largest number of query
heads per layer ($32$) but the \emph{smallest} GQA sharing ratio
of the three.

\subsection{Datasets}
\label{sec:datasets}

\textbf{Main (Exp.\,1--2):} 500 texts from the \texttt{allenai/c4}
English validation split, streamed and shuffled with seed 42
(buffer 10,000). C4 shares CommonCrawl provenance with
Pile-CC and serves as a practical substitute
here~\cite{raffel2020c4}. Two sampling procedures are used.
At 2,048 and 4,096 tokens, documents are tokenized with
\texttt{add\_special\_tokens=False}, appended to a rolling token
buffer, cut into non-overlapping windows of exactly the target
length and decoded back to text, so every sample fills the
context window; no separator and no BOS token are inserted
between documents. Windows are built with the Qwen2.5-0.5B
tokenizer, and each model re-tokenizes with its own at inference
time, so realized token counts differ slightly across models.
At 512 tokens there is no concatenation: whole documents passing
a 200-word minimum filter are truncated to 512 tokens, and
shorter documents are left short. The 512-token condition
therefore has a lower and more variable effective length than the
other two --- an asymmetry between conditions, not a controlled
manipulation.

\textbf{Calibration transfer probe (Exp.\,4):} LongBench v1
\texttt{multifieldqa\_en}~\cite{bai2024longbench}, 75 examples.

\section{Results}

\subsection{Main STC Results (Experiment 1)}

Table~\ref{tab:main} reports the full STC family at 4,096 tokens.
All three models show the same Rank--Set contrast: global
Cal-STC-Rank is at least $0.980$, but Cal-STC-Set is
$0.619$--$0.793$. Because Set@$k$ is Jaccard overlap, these values
mean that $76.5$--$88.5\%$ of the selected heads are retained after
quantization. This is far above the null baseline ($0.053$),
yet still consequential for a policy that treats top-$k$ membership
as binary.
The top-$k$ heads exhibit mean bf16 SCS of $0.65$--$0.80$
($2{,}660$--$3{,}260{\times}$ the uniform-attention value
$1/4096$), confirming genuine first-token attention concentration.

\begin{figure}[t]
\centering
\includegraphics[width=0.97\columnwidth]{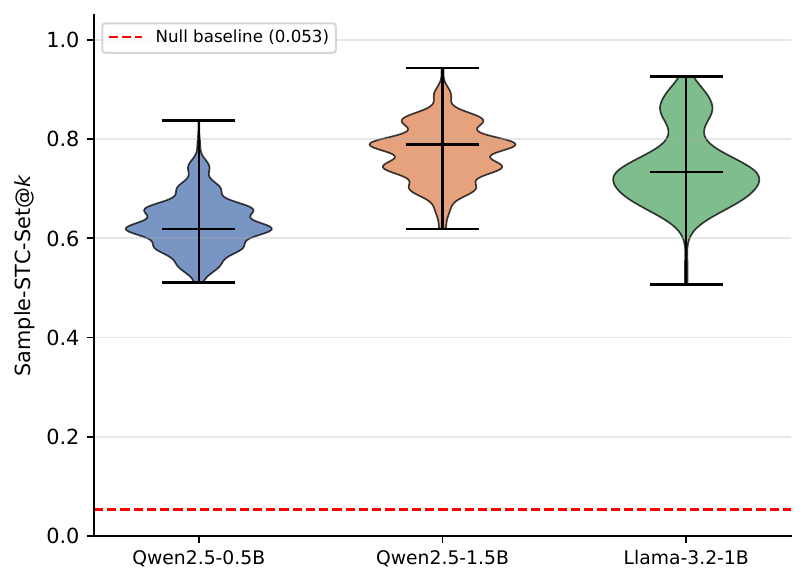}
\caption{Sample-STC-Set@$k$ across the 500 inputs (violin). The
dashed line is the null baseline ($0.053$); all three
distributions sit far above it.}
\label{fig:violin}
\end{figure}

\begin{table}[ht]
\caption{STC Metric Family at 4,096 Tokens
(Fixed-length C4 windows, 500 samples).
Samp Set@$k$ is mean $\pm$ 1 SD across the 500 inputs, not a
confidence interval. Null is the closed form $k/(2LH_q-k)$.}
\label{tab:main}
\centering
\setlength{\tabcolsep}{4pt}
\begin{tabular}{lccccc}
\toprule
Model & Cal & Cal & Cal & Samp & Null \\
      & Rank & Set@$k$ & LS$\downarrow$ & Set@$k$ & @$k$ \\
\midrule
Qwen2.5-0.5B & 0.980 & 0.619 & 0.045
  & $0.629\pm0.053$ & 0.053 \\
Qwen2.5-1.5B & 0.990 & 0.789 & 0.029
  & $0.771\pm0.057$ & 0.053 \\
Llama-3.2-1B & 0.983 & 0.793 & 0.019
  & $0.745\pm0.076$ & 0.053 \\
\bottomrule
\end{tabular}
\end{table}

The contrast has a simple operational interpretation: a high global
rank correlation does not guarantee a stable discrete selection.
Small score changes near the $k$-th boundary can exchange heads while
leaving the full ordering highly correlated. The two summary metrics
do not identify the perturbation mechanism, and the global rank can
also be dominated by between-layer score differences
(Section~\ref{sec:layerwise}). Thus the supported guidance is
narrower than ``rank is robust'': reuse a global continuous ranking
with caution, and revalidate any binary or layer-local head policy.

Llama-3.2-1B attains the highest Cal-Set ($0.793$) together with the
lowest LayerShift ($0.019$); Qwen2.5-0.5B has the lowest Cal-Set
($0.619$). With only three models, this ordering supports neither a
size nor a family explanation (Section~\ref{sec:layerwise}).

\subsection{Across Context Lengths (Experiment 2)}

\begin{table}[ht]
\caption{Cal-STC-Set@$k$ Across Context Lengths. Packed
fixed-length windows at 2,048 and 4,096; truncated whole documents
at 512 (Section~\ref{sec:datasets}).}
\label{tab:context}
\centering
\begin{tabular}{lccc}
\toprule
Model & 512 & 2,048 & 4,096 \\
\midrule
Qwen2.5-0.5B & 0.744 & 0.744 & 0.619 \\
Qwen2.5-1.5B & 0.700 & 0.744 & 0.789 \\
Llama-3.2-1B & 0.705 & 0.763 & 0.793 \\
\bottomrule
\end{tabular}
\end{table}

Qwen2.5-0.5B declines from 0.744 to 0.619 at 4,096 tokens, while
Qwen2.5-1.5B and Llama-3.2-1B increase
(0.700$\to$0.789 and 0.705$\to$0.793).
All values remain substantially above the null
baseline ($0.053$) at all tested lengths.
Because the 512-token condition uses truncated whole documents rather
than packed windows, the table is a robustness check across operating
points, not a controlled estimate of context-length causality.

\subsection{Layerwise Analysis (Experiment 3)}
\label{sec:layerwise}

Layerwise analysis localizes where sink mass moves under
quantization. It does not by itself locate the cause of global set
instability: Section~\ref{sec:recal} shows that updating a broader,
preselected group of high-shift layers still does not restore the
top-$k$ set. The terminal spike and the global Rank--Set contrast are
therefore related observations, not one explanation for the other.

\begin{figure}[t]
\centering
\includegraphics[width=0.97\columnwidth]{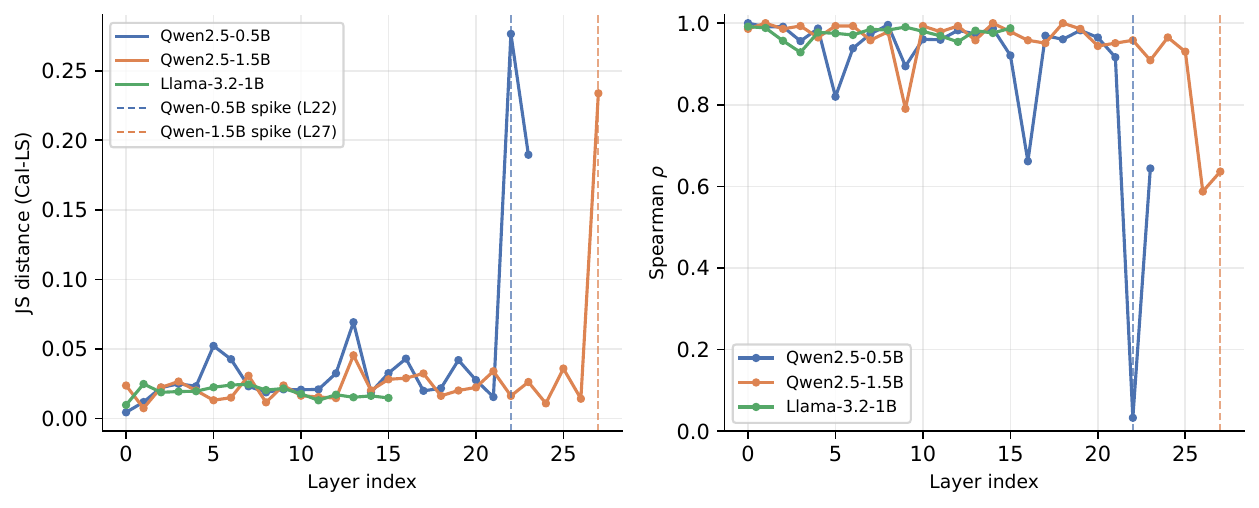}
\caption{Per-layer sink topology shift under 4-bit PTQ.
Left: LayerShift (JS distance) per layer. Right: Spearman
rank correlation per layer. Model-colored dashed vertical lines mark
the peak-shift layers for Qwen2.5-0.5B (layer 22)
and Qwen2.5-1.5B (layer 27). Llama-3.2-1B shows uniform,
low-magnitude drift throughout.}
\label{fig:layerwise}
\end{figure}

\begin{table}[ht]
\caption{Layerwise Sink Topology Shift
(Exp.\,3, 4,096 tokens). ``Peak layer'' is the single layer of
maximum LayerShift; JS and Rank are that layer's values.
Global mean LayerShift:
Qwen2.5-0.5B\,=\,0.045,
Qwen2.5-1.5B\,=\,0.029,
Llama-3.2-1B\,=\,0.019. Ratios in the text use unrounded values.}
\label{tab:layerwise}
\centering
\begin{tabular}{lcccl}
\toprule
Model & Peak layer & JS & Rank & Pattern \\
\midrule
Qwen2.5-0.5B & 22 (last-2) & 0.276 & 0.033
  & End-layer spike \\
Qwen2.5-1.5B & 27 (last)   & 0.234 & 0.636
  & End-layer spike \\
Llama-3.2-1B & 1 (early)   & 0.025 & 0.988
  & Uniform, flat \\
\bottomrule
\end{tabular}
\end{table}

High-shift layers are defined as those whose LayerShift exceeds
$3\times$ the model mean. This yields layers 22--23 for
Qwen2.5-0.5B ($6.2\times$ and $4.2\times$ the mean) and layer 27
alone for Qwen2.5-1.5B ($7.9\times$). For Llama-3.2-1B no layer
qualifies; its maximum is $1.3\times$ the mean. The peak layer
thus carries $6.2\times$ (Qwen2.5-0.5B) and $7.9\times$
(Qwen2.5-1.5B) the model-mean shift. Rank correlation at these
layers falls to $0.033$ (layer 22), $0.644$ (layer 23) and
$0.636$ (layer 27); Qwen2.5-0.5B layer 22 is close to complete
rank collapse, meaning the sink ordering at that layer is
essentially uninformative after 4-bit quantization.

Rank degradation is not confined to the high-shift layers, which
qualifies the Rank--Set dissociation. Excluding the high-shift layers, the
lowest per-layer Rank is $0.662$ (Qwen2.5-0.5B, layer 16),
$0.587$ (Qwen2.5-1.5B, layer 26) and $0.929$ (Llama-3.2-1B,
layer 3). Qwen2.5-1.5B layer 26 is the clearest case: its
LayerShift ($0.014$) is below the model mean, yet its Rank is the
lowest in the model. Per-layer Rank and LayerShift thus capture
partly independent failure modes, and the near-unity
\emph{global} Cal-STC-Rank of Table~\ref{tab:main} reflects the
large between-layer spread in sink mass rather than uniformly
preserved within-layer ordering.

The mechanism of the Qwen end-layer spike remains open. The terminal
layers feed the LM head directly and may be more sensitive to
weight perturbation under NF4, but per-layer weight dynamic range
($\max-\min$ over the layer's attention and FFN projection
weights) does not predict LayerShift in any model: Spearman
$\rho = 0.05$ ($p{=}0.81$), $0.01$ ($p{=}0.96$) and $-0.36$
($p{=}0.18$) for Qwen2.5-0.5B, Qwen2.5-1.5B and Llama-3.2-1B.
With only $L = 24, 28, 16$ points, these correlations do not support
dynamic range as an explanation; they do not rule it out. The GQA
ratios likewise oppose the simplest sharing account: Llama-3.2-1B
has the \emph{smallest} query-to-KV ratio ($4$ versus $7$ and $6$)
despite the mildest layerwise shift. We therefore leave the observed
model-family contrast unexplained.

\subsection{Cross-Domain Calibration Transfer (Experiment 4)}

\begin{table}[ht]
\caption{Calibration Transfer to LongBench (75 examples, 4,096
tokens). $A$: LB bf16 vs.\ LB 4-bit; $B$: C4 bf16 vs.\ LB 4-bit;
$C$: C4 bf16 vs.\ LB bf16. Contrasts are descriptive and computed
from unrounded values.}
\label{tab:transfer}
\centering
\setlength{\tabcolsep}{3pt}
\begin{tabular}{lccccc}
\toprule
Model & $A$ & $B$ & $C$ & Corpus & Prec. \\
      & in-dom & deploy & dom-only & $\Delta$ & $\Delta$ \\
\midrule
Qwen2.5-0.5B & 0.659 & 0.447 & 0.545 & $+$0.212 & $+$0.099 \\
Qwen2.5-1.5B & 0.889 & 0.360 & 0.360 & $+$0.529 & $+$0.000 \\
Llama-3.2-1B & 0.733 & 0.793 & 0.825 & $-$0.060 & $+$0.032 \\
\bottomrule
\end{tabular}
\end{table}

Table~\ref{tab:transfer} compares deployment overlap $B$ with two
controlled references. The corpus contrast ($A-B$) changes the
reference map while fixing the 4-bit target; the precision contrast
($C-B$) changes target precision while fixing the C4 bf16 reference.
Because overlap is non-additive, these contrasts reveal which
comparison changes more; they do not assign independent causal shares.

Llama-3.2-1B shows a negative corpus contrast ($-0.060$):
the C4-calibrated map overlaps \emph{more} with the 4-bit
LongBench deployment map than the in-domain bf16 map does. Its
precision contrast is also small ($+0.032$). Among the tested models,
its calibration map is therefore least sensitive to this particular
corpus-and-precision shift.

For both Qwen models the corpus contrast exceeds the precision
contrast, but by different margins. For Qwen2.5-1.5B, $B$ and $C$
have equal overlap ($0.360$) with the C4 reference, so this comparison
shows no additional scalar overlap loss from target precision, while
the corpus contrast is $+0.529$. Equality of the scalar does not imply
identical LongBench head maps. For Qwen2.5-0.5B, the precision contrast
is $+0.099$ against a corpus contrast of $+0.212$; both are material.

The deployment lesson is that precision-only STC is insufficient:
corpus mismatch can reduce overlap even more. A sink map intended for
a new domain should therefore be checked on representative target
inputs, especially for the two Qwen models studied here.

\subsection{Recalibration Sample Efficiency (Experiment 5)}
\label{sec:recal}

\begin{figure}[t]
\centering
\includegraphics[width=0.97\columnwidth]{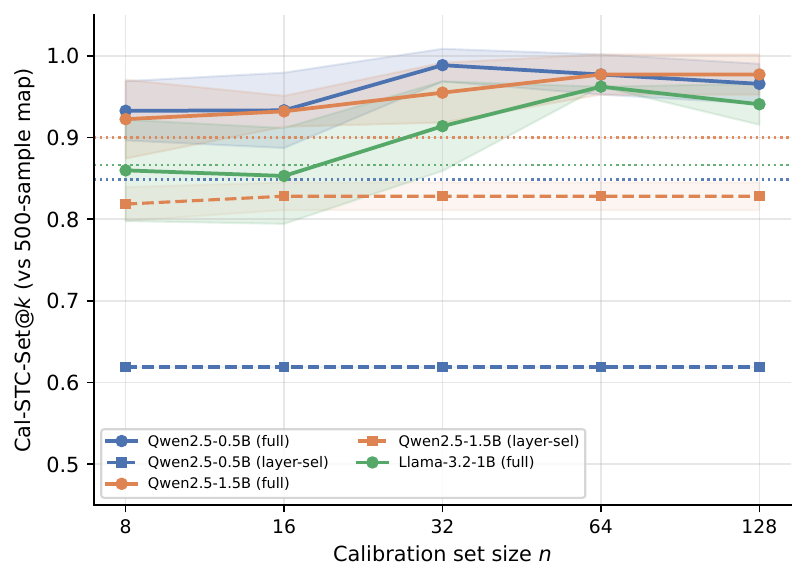}
\caption{Set@$k$ agreement between the $n$-sample 4-bit
calibration map and the 500-sample 4-bit map. Solid lines show
full recalibration; dashed lines show layer-selective recalibration
for the two Qwen models (undefined for Llama-3.2-1B). Model-colored
dotted lines mark 90\% of each model's split-half plateau
(0.849, 0.900 and 0.866). Shaded bands are normal-approximation
95\% confidence intervals from five subsamples.}
\label{fig:convergence}
\end{figure}

\begin{table}[ht]
\caption{Samples Needed to Reach 90\% of Split-Half Stability.
Candidate $n\in\{8,16,32,64,128,500\}$.}
\label{tab:recal}
\centering
\begin{tabular}{lcccc}
\toprule
Model & Plateau & Thresh. & Full recal & Layer-sel \\
      &         &         & suff.\ $n$ & suff.\ $n$ \\
\midrule
Qwen2.5-0.5B & 0.943 & 0.849 & 8  & $\dagger$ \\
Qwen2.5-1.5B & 1.000 & 0.900 & 8  & $\dagger$ \\
Llama-3.2-1B & 0.962 & 0.866 & 32 & $\ddagger$ \\
\bottomrule
\end{tabular}
\par\vspace{2pt}
{\raggedright\small $\dagger$Never reaches the threshold: saturates at $0.619$
and $0.838$ regardless of $n$. $\ddagger$Undefined ---
Llama-3.2-1B has no high-shift layer, so the implementation falls
back to the full map and the two columns would coincide by
construction, not by measurement.\par}
\end{table}

This experiment reuses the 4-bit per-sample maps from Experiment 1 at
4,096 tokens; it is a convergence study within the C4 pool, not a new
quantization experiment or an independent-domain generalization test.
The reference is the 500-sample 4-bit map. Because this reference and
the $n$-sample estimates share a finite pool, we use agreement between
two disjoint 250-sample maps as an empirical plateau and call $n$
sufficient at 90\% of it. For each $n$ we draw five subsamples from a
single \texttt{numpy.random.default\_rng(42)} stream and report their
mean and standard deviation.

Layer-selective recalibration substitutes the $n$-sample 4-bit
estimate into the high-shift layers only, leaving the bf16 map
elsewhere. The sets used, $\{22,23\}$ and $\{25,26,27\}$, were
fixed beforehand by a top-quartile LayerShift criterion and are
broader than the $3\times$-mean sets of
Section~\ref{sec:layerwise}; even these broader updates fail to reach
the threshold. The result shows that these selected layers alone do
not recover the global top-$k$ set. Full-map estimates reach the
criterion at the smallest tested $n{=}8$ for both Qwen models.
For Qwen2.5-0.5B, replacing layers 22--23 changes their scores but
no global top-$k$ membership even at $n{=}500$; its selected set
therefore remains the bf16 set, and the $0.619$ plateau exactly
matches Cal-STC-Set in Table~\ref{tab:main}. For Qwen2.5-1.5B, the
$n{=}500$ update changes one top-$k$ member and raises agreement from
$0.789$ to $0.838$.

For Llama-3.2-1B the smallest sufficient $n$ is 32, but not as a
sharp threshold: $n{=}8$ reaches $0.860$ against a threshold of
$0.866$, a shortfall far below the across-subsample SD there
($0.071$), and the curve is non-monotonic from $n{=}8$ to
$n{=}16$. With five subsamples the supportable claim is that
Llama-3.2-1B needs more calibration data than the Qwen models,
not exactly four times as much. Its per-sample Set@$k$
variability is likewise the highest of the three
($\mathrm{SD}=0.076$, Table~\ref{tab:main}), consistent with a
noisier top-$k$ boundary. We did not isolate its cause; notably,
Llama-3.2-1B has the smallest GQA sharing ratio in this model set.

\section{Jetson Orin NX Deployment Validation}

We measure deployment cost on an NVIDIA Jetson Orin NX 8GB using the
same HuggingFace Transformers and bitsandbytes NF4 code path at the
library versions available for arm64. Table~\ref{tab:jetson} reports
prefill latency (TTFT), peak GPU memory, hook overhead, and elapsed
time for a 16-sample recalibration workload. Valid map recovery is
reported separately because one model encounters a software issue.

\begin{table}[ht]
\caption{Jetson Orin NX 8GB Deployment Results (HF + bitsandbytes
NF4). Rec16 is elapsed time for the 16-sample workload.}
\label{tab:jetson}
\centering
\setlength{\tabcolsep}{3pt}
\begin{tabular}{lcccccc}
\toprule
Model & \multicolumn{2}{c}{TTFT (ms)} & Peak & Hook & Hook & Rec16 \\
      & @512 & @4k & GPU(MB) & (ms) & (\%) & (s) \\
\midrule
Qwen2.5-0.5B & 176 & 274 & 926  & 8.3  & 4.8 & 2.9 \\
Llama-3.2-1B & 353 & 444 & 1513 & 19.5 & 5.5 & 5.7 \\
Qwen2.5-1.5B & 400 & 535 & 1634 & 37.2$^\dag$ & 9.3$^\dag$ & 6.8$^\dag$ \\
\bottomrule
\end{tabular}
\end{table}
{\small \dag For Qwen2.5-1.5B the hook \emph{values} are invalid on
this platform: attention weights returned NaN under bitsandbytes
NF4 + eager attention (transformers 4.57.6 on arm64), a
version-specific compatibility issue, so no sink map could be
recovered on-device. The timings remain valid --- the hook performs
the same tensor operations whether or not the values are finite.
No STC result here depends on the Jetson hook output.
Peak GPU(MB) is \texttt{torch.cuda.max\_memory\_allocated()};
Jetson Orin NX uses unified memory.}

All three models fit within the 8\,GB unified memory budget, with
TTFT from 176 to 535\,ms. Hook-based SCS accumulation adds
8.3--37.2\,ms per forward pass (4.8--9.3\% of base TTFT), and the
16-sample workload completes in 2.9--6.8\,s. These measurements
validate seconds-scale map recovery for Qwen2.5-0.5B and
Llama-3.2-1B. For Qwen2.5-1.5B, they validate resource cost only;
the tested arm64 stack returns invalid attention values, so on-device
map recovery remains unvalidated for that model.

\section{Discussion}

\textbf{Global stability is not local stability.}
Quantization retains $76.5$--$88.5\%$ of top-$k$ membership while
global rank correlation stays above $0.980$. The apparent tension is
the paper's central result: a continuous global summary can remain
stable while a thresholded decision changes. Moreover, the global
rank partly reflects between-layer score separation; it does not
certify within-layer head order.

\textbf{End-layer spike as an architectural marker.}
The Qwen terminal-layer spike is a marker, not an explanation.
Neither per-layer weight dynamic range nor the simple ordering of GQA
sharing ratios supports a mechanism in this three-model study, and
targeting a broader set of high-shift layers does not restore the
global top-$k$ set.

\textbf{Corpus and precision are separate checks.}
For both Qwen models, changing from matched-domain to cross-domain
calibration reduces overlap more than changing the target from bf16
to 4-bit under the tested C4-to-LongBench shift. Llama-3.2-1B is less
sensitive to both comparisons. These non-additive contrasts do not
identify causal shares, but they show why a precision-only check can
miss the larger deployment mismatch.

\textbf{Deployment rule.}
Reuse a bf16 global ranking only when the downstream method consumes
that ranking globally. Recheck discrete top-$k$ sets, layer-local
policies, and new domains in 4-bit. In the C4 resampling study, full
maps reach the stability criterion with the smallest tested
$n{=}8$--$32$; for the two Qwen models, selected-layer updates do not
reach the criterion. Jetson measurements show seconds-scale recovery
for the two models with valid hooks, but not yet for Qwen2.5-1.5B on
the tested arm64 software stack.

\textbf{Limitations.} Three models from two families cannot separate
family, architecture, and size. Results cover bitsandbytes NF4
(fp16 compute versus a bf16 baseline; no double quantization), one
cross-domain pair, and a finite-pool resampling study. STC measures
topology, not task impact: we do not show that map churn degrades
end-task accuracy. The guidance therefore concerns calibration
validity, not application quality.

\balance
\section{Conclusion}

Four-bit NF4 preserves global BOS-sink head order better than it
preserves the selected head set. Across three small language models,
global rank remains at least $0.980$, yet only $76.5$--$88.5\%$ of
top-$k$ membership is retained, and terminal Qwen layers exhibit
sharp local shifts hidden by the global statistic. Cross-domain
calibration can reduce overlap more than precision alone. A practical
pipeline should therefore reuse continuous global rankings
selectively, but revalidate binary sets, layer-local policies, and
new domains in the deployed precision. In our C4 study, full 4-bit
maps reach an empirical stability criterion with the smallest tested
$n{=}8$--$32$, and valid Jetson measurements show that this check can
run in seconds.

{\renewcommand{\baselinestretch}{0.97}\footnotesize
\bibliographystyle{IEEEtran}

}

\end{document}